**Early Prediction of Alzheimer's Disease Leveraging Symptom Occurrences from Longitudinal Electronic Health Records of US Military Veterans**

Rumeng Li[1,2], MS; Xun Wang[2], PhD; Dan Berlowitz[3,4], MD; Brian Silver[5], MD; Wen Hu[2,6], MS; Heather Keating[2,6], PhD; Raelene Goodwin[2,6], BS; Weisong Liu[2,4,6], PhD; Honghuang Lin[7], PhD; Hong Yu[1,2,4,6], PhD

[1]Manning College of Information & Computer Sciences, University of Massachusetts Amherst, Amherst, MA, USA;

[2]Center for Healthcare Organization & Implementation Research, VA Bedford Health Care System, Bedford, MA, USA;

[3]Department of Public Health, University of Massachusetts Lowell, Lowell, MA, USA;

[4]Center of Biomedical and Health Research in Data Sciences, University of Massachusetts Lowell, Lowell, MA, USA;

[5]Department of Neurology, UMass Chan Medical School, Worcester, MA, USA

[6]Miner School of Computer & Information Sciences, University of Massachusetts Lowell, Lowell, MA, USA;

[7]Department of Medicine, UMass Chan Medical School, Worcester, MA, USA

alicerumengli@gmail.com/ Hong_Yu@uml.edu

**Abstract**

**Background and Objectives:** Early prediction of Alzheimer's disease (AD) is crucial for timely intervention and treatment. This study aims to use machine learning approaches to analyze longitudinal electronic health records (EHRs) of patients with AD and identify signs and symptoms that can predict AD onset earlier.

**Methods:** We used a case-control design with longitudinal EHRs from the U.S. Department of Veterans Affairs Veterans Health Administration (VHA) from 2004 to 2021. Cases were VHA patients with AD diagnosed after 1/1/2016 based on ICD-10-CM codes, matched 1:9 with controls by age, sex and clinical utilization with replacement. We used a panel of AD-related keywords and their occurrences over time in a patient's longitudinal EHRs as predictors for AD prediction with four machine learning models. We performed subgroup analyses by age, sex, and race/ethnicity, and validated the model in a hold-out and "unseen" VHA stations group. Model discrimination, calibration, and other relevant metrics were reported for predictions up to ten years before ICD-based diagnosis.

**Results:** The study population included 16,701 cases (488 [2.9%] women; mean [SD] age, 76.7 [8.6] years) and 39,097 matched controls (1,013 [2.6%] women; mean [SD] age, 76.6 [8.9] years). The average number of AD-related keywords (e.g., "concentration", "speaking") per year increased rapidly for cases as diagnosis approached, from around 10 to over 40, while remaining flat at 10 for controls. The best model achieved high discriminative accuracy (ROCAUC 0.997) for predictions using data from at least ten years before ICD-based diagnoses. The performance of using ICD codes only was poor (ROCAUC 0.536 and ROCAUC 0.500 using data 1 and 10 years before diagnosis). The model was well-calibrated (Hosmer-Lemeshow goodness-of-fit p-value = 0.99) and consistent across subgroups of age, sex and race/ethnicity, except for patients younger than 65 (ROCAUC 0.746).

**Discussion:** AD-related signs and symptoms are reported in EHRs many years before clinical diagnosis and their frequency, approximated by AD-related keywords, increases the closer one is to the diagnosis. Machine learning models using AD-related keywords identified from EHR notes can predict future AD diagnoses, suggesting its potential use for identifying AD risk solely using EHR notes, offering an affordable way for early screening on large population.

## Introduction

Alzheimer's disease (AD) is a progressive neurodegenerative disorder resulting in impairment of memory and other cognitive functions.[1] The number of people suffering AD could rise from approximately five million today to between 11 and 16 million in the United States, and more than 100 million worldwide by 2050.[2] Early prediction of AD risk is a key to effective interventions and treatments,[1,2] potentially leading to new therapies that alter the course of cognitive decline.

Existing research on AD diagnosis generally focuses on AD biomarkers of functional impairment, neuronal loss, and abnormal protein aggregation noted on neuroimaging, cerebrospinal fluid analysis, and serum evaluation.[3–5] However, these tests are often time consuming and expensive and have not been widely adopted in the clinical practice. Yet AD often causes or is accompanied by other physical and mental health changes[2] that are noticeable and often captured in routine health care services. These changes have been leveraged to detect AD using patient speech records and other clinic data.[6–17] However, these studies have been limited by: 1) focusing on one aspect of AD symptoms or risk factors rather than multiple ones[6–8] and 2) relying on diagnostic codes, medications, procedure codes for psychological/cognitive testing without taking advantage of the clinical information in EHRs to make AD risk predictions,[9,10,14–17] as subtle changes may not be recorded as ICD codes and reflected by prescriptions but recorded as plain texts in EHRs.

Prior work has shown that biological changes and accompanying symptoms can happen long before formal AD diagnosis. Abnormal buildups of proteins that form amyloid plaques and tau tangles were observed decades before AD onset.[18] Reports of memory loss occur up to 12 years prior to AD diagnosis as do depressive symptoms.[19] Poor visual memory is associated with an increased risk of AD up to 15 years later.[20] Pre-diagnostic cognitive impairment and decline appear nine years before AD diagnosis.[21] These signs and symptoms are usually described in the unstructured EHR notes, and may be difficult to distinguish from normal aging.[2] Leveraging these clues may be helpful to the early detection of AD risk.[22,23] Manually reviewing years of EHR notes is costly and non-scalable. Therefore, in this work we investigate machine learning approaches to address the following three questions. First, what early signs and symptoms appearing in EHRs may predict AD? Second, when do possible early signs and symptoms begin to increase in frequency in the EHRs for AD patients? Finally, could these signs and symptoms help identify populations at high risk of AD?

## Methods

### Setting and Data

This study used de-identified longitudinal EHR data from patients receiving care in the US Department of Veterans Affairs Veteran Health Administration (VHA) from January 1, 2004 to October 1, 2021. The study was approved by the Institutional Review Board at the VHA Bedford Healthcare System, which also approved the waiver of documentation of informed consent.

### Study Design

An incidence-based, case-control design was used in this study. Figure 1 shows the inclusion-exclusion criteria for generating the cohort. The cases are patients with first AD diagnosis between January 1, 2016 and October 1, 2021 based on ICD codes (eTable 1 in the Supplement). To improve the accuracy of AD case ascertainment, an AD case must be either diagnosed once at a specialty dementia clinic in VHA or diagnosed at least twice at separate time points with at least one diagnosis generated at specialty clinics including neurology, geriatrics, GeriPACT (Geriatric Patient Aligned Care Team), mental health, psychiatry, psychology, or geriatric psychiatry with a provider type restricted to neurology, vascular neurology, psychiatry, neuropsychology and geriatric medicine. The clinic types are identified by Stop Codes (eTable 2 in the Supplement) which are used by VHA to specify the type of outpatient care and workload of a visit.[24]

AD index date was defined as the first ICD-based AD diagnosis date for a patient. AD patients with less than five years continuous EHR data prior to one's AD index date were excluded. We matched nine controls using non-AD patients with replacement by age (±1 year), sex, and clinical utilization (Cases and controls had their first visit in the same year within the study period and at least one visit in the year of the case diagnosis.). We included patients younger than age 65 because early-onset AD comprises about 5% to 6% of cases of AD.[25] By design, a patient could be a control participant for multiple case participants, a case participant before diagnosed could not serve as a control participant for another case participant.

There are hundreds of different note types in the EHRs. In this study, we only included EHRs generated by primary care visits, memory clinics, neurology, geriatric psychiatry, geriatric medicine, nurses providing cognitive care, mental health clinics, compensation and pension examination[26] and consultation visits. Of which, primary care notes comprised 41% of all notes for cases and 56% for controls. Notes from the six sources (primary care, geriatric psychiatry, geriatric medicine, consultation, compensation and pension examination) made up 68% of all notes for cases and 81% for controls. Supplemental eFigure 1(a) shows the note type distribution in the cohort and Figure 1(b) shows the average note number per patient by note type.

**Predictors**

Predictors studied in this work are the occurrences of age-specific AD-relevant keywords. Specifically, through iterative literature review, domain experts curated a widely inclusive list of 131 AD-relevant keywords (Supplemental eBox 1). These keywords related to cognition (memory, learning, language), physiology, mood, as well as cognitive and diagnostic testing. For each patient we identified the occurrence of keywords in longitudinal EHRs and aligned them to the patient's age $age_j$ to form a temporal representation for each keyword $k_i$ as $\{< age_j, k_i >\}$, allowing easy tracking of each keyword with increasing ages. Figure 2 provides illustrative examples on predictor extraction. The patient's records state "good mood" at $age_1$ and "mood dysphoric" at $age_3$; the keyword "mood" is extracted for both without differentiating the contextual meanings. The keywords reflect clinicians' noting/documenting a feature of the patient. If the feature is concerning, it may be noted more frequently. We examined the frequency of these keywords approximating AD-related signs and symptoms. We used term frequency-inverse document frequency (TF-IDF) for identifying the importance of each $\{< age_j, k_i >\}$(eMethods in the Supplement). We identified 8,751 unique keyword-age pairs. Since the majority of pairs appear sparsely, we kept the most frequent 1000 keyword-age pairs as going beyond this point led to little improvement for overall performances. These 1,000 keyword-age pairs (eBox 2 in the Supplement) are the predictors in this study thus we considered both the occurrence of a keyword and the frequency with which it appears over time.

**Machine Learning Models for Predicting AD**

Prediction of a patient at risk of AD can be formulated as a classification task. The inputs to the machine learning models are the predictors of a patient (vector representation) and outputs of the models are the risk of AD. We deployed machine learning models widely used for predictions in the clinical domain,[27] including logistic regression (LR),[28] support vector machine (SVM), [28] AdaBoost[28] and random forests (RF)[28] (eMethods in the Supplement).

We conducted experiments under two settings. In Setting I, we randomly split the study cohort into 80% (training) and 20% (testing). In Setting II, we selected patients from 10 randomly chosen VHA stations (out of 130) for testing and the remaining are used for training (92%). Patients with multiple stations visited were assigned to the most frequently visited station.

### Observation Window and Prediction window

For each setting, the observation window spanned from the start of the study period on 1/1/2004 to the start of a clean period in which no data is used in predicting AD. The length of this clean window could vary from 0 to 10 years where a clean window of 0 years means that data from up to the day before the diagnosis of AD is used in prediction (Figure 3). Those with less than 5 years EHR history within the observation window were excluded. Predictors were extracted from the observation window. When the observation window concluded at the AD index date, there were a total of 55,798 individuals divided between the training and testing sets (eTable 3 in the Supplement). As the prediction period (clean window) was extended from 0 to 10 years, the number of people included in the test set decreased.

### Evaluation Metrics and Error Analyses

For performance evaluation, we adopted the following criteria: the precision, recall, F1 score, accuracy, the area under the receiver operating characteristic curve (ROC AUC) and the area under the precision-recall curve (PR AUC), with a predictive probability threshold of 0.5 set empirically. We conducted calibration analysis based on the LR model. The calibration was assessed using Hosmer-Lemeshow goodness-of-fit test[29] with five sub-groups and was graphically depicted in calibration plots. A p-value greater than 0.05 indicates sufficient goodness-of-fit. We also calculated Shapley values[30] to identify the key predictors contributed to the prediction. Details of the evaluation metrics are in eMethods in the Supplement.

For error analyses, domain experts (BS and WH) conducted chart review on incorrect predictions.

### AD-related Keyword Occurrence Patterns in Longitudinal EHRs

We identified AD signs and symptoms in longitudinal EHRs using expert-curated keywords (Supplemental Box 1). We computed the average number of AD-related keywords per year for up to 15 years before AD diagnosis for the whole cohort and by age group. We examined the patterns using all the notes for our selected note types and only primary care notes. Since longitudinal analysis using all the notes of patients is costly, we randomly sampled 100 cases and 100 controls from our cohort and analyzed the keyword occurrence pattern in all of their EHR notes. In addition, we expected that patients with AD symptoms would have more frequent physician visit than those without such symptoms, and thus the AD cases would have more notes with AD-related keywords than the controls. To account for this potential bias, we normalized the AD keyword frequency by note number.

**Sensitivity Analysis**

To further test the robustness of our model, we conducted a series of sensitivity analyses. First, we stratified the cohort by sex, age, and race to test the sensitivity of the predictive model to biological variables. Second, as previous work mostly used structured EHR data for prediction,[9,10,14–17] we also compared the performance of our prediction model by using different predictors: ICD codes only, structured data only (demographics, ICD codes and medications), and both structured data and keywords. Implementation details are in eMethods in the Supplement. Third, we excluded cognitive test keywords (e.g., “MMSE” and “Mini-Cog”) from the predictors because such tests are usually used to assess the extent of memory impairment and indicate the clinicians’ suspicion or awareness of the patients’ AD risk. Therefore, we only used keywords related to AD signs and symptoms identified through routine clinical care records for AD risk prediction. Fourth, we used only primary care notes for model training and testing as the primary care setting is often a patient’s entry point into the healthcare system.[31] This approach intended to address the potential bias from specialty clinics that are specific to AD concerns and might indicate early AD risk. Fifth, as AD diagnosis is challenging[32], existing work also employed different criteria for identifying AD cases,[15] we tested the model on different groups of AD patients based on their diagnosis codes. These groups were: cases with a single AD diagnosis at dementia clinics, cases with multiple AD diagnoses including at least one from specialty clinics, and AD patients with a single AD diagnosis from any clinics or with two or more diagnoses from non-specialty clinics. We also checked the performance by stratifying the cases with multiple diagnoses by the time interval between the first and second diagnosis.

**Data Availability**

The data used in the preparation of this Article are from VHA. Approval by the Department of Veterans Affairs is required for data access.

## Results

Our case cohort comprised 16,701 AD patients (488 [2.9%] women; mean [SD] age, 76.7 [8.6] years) and the control cohort comprised 39,097 patients (1,013 [2.6%] women; mean [SD] age, 76.6 [8.9] years). All ages were as of 1/1/2016. White patients accounted for 77.4% in the case patients and 79.9% in the control. For Black or African American, the ratios are 14.1% and 10.3% respectively. Hispanic or Latino ethnicity accounted for 11.2% in the case patients and 4.5% in the control group. Detailed characteristics of the study cohort are in Table 1. Among AD cases (16,701 patients), the majority of 16,404 patients (98.2%) had multiple ICD codes for AD with at least one diagnosis code from specialty clinics and the remaining 297 patients (1.8%) had a single dementia clinic-based diagnosis.

According to the Shapley value, the most important predictors are the following keywords combined with different ages including "MMSE", "concentration" and "speaking" (eBox 2 in the Supplement). We further grouped the keywords into eight subgroups: Cognition-Speech/Language, Cognition-Memory, Cognition-Other, Physiology/Behavior Changes, Mood, Testing, Anatomy and Others (eBox 3 in the Supplement). In general Cognition-Speech, Testing, and Cognition-Memory keywords are the most important predictor groups. The importance of predictors for AD changes over people's ages. The most important predictor group for patients age 60 and below is Cognition-Memory, Testing and Cognition-Speech for patients in 60s, Cognition-Speech and Cognition-Other for patients in 70s, and Testing for age 80 and higher (eFigure 2 in the Supplement).

The average number of AD-related keywords per year in the notes of AD patients started to deviate from those of controls about 14 years before the AD diagnosis (Figure 4a). This pattern was consistent across different ages at diagnosis (Figure 4b). We replicated this finding using only primary care notes (eFigure 3 in the Supplement), a random subset of cases and controls using all their EHR notes (eFigure 4 in the Supplement), and a normalized keyword frequency by notes number (eFigure 5 in the Supplement).

We constructed predictive models on the two training/testing settings and the results showed little difference (Table 2 and eTable 4(a), 4(b) in the Supplement). Therefore, we mainly report the results of Setting I. The results of making predictions using data from only 10 or more years prior to AD diagnosis are shown in Table 2. Predicting AD diagnosis 10 years in advance requires a patient to have at least 15 years EHR history, which resulted in a test set of 4076 patients (AD cases: 1112). The LR model reported an accuracy 0.96. SVM had similar predictions with accuracy 0.95 while others (AdaBoost and RF) performed significantly lower (accuracy 0.89 and 0.81, respectively). Results continue to improve when making predictions using data more recent than 10 years (eTable 5 in the Supplement). For making predictions at other time points ($N-k, k=0,1,2 \ldots 9$), patients' note distribution and patients' demographic distribution in different test sets and the results of LR are shown in eTable 5 in the Supplement. The highest F1 score of 0·99 and accuracy of 1 is achieved making prediction in the same year of AD diagnosis. The model performance decreases over time but still maintains F1 score of 0·93 and accuracy of 0·96 when making predictions ten years prior to AD diagnosis.

The calibration plot approaches linearity (eFigure 6 in the Supplement), with Hosmer-Lemeshow goodness-of-fit p-values=0.99 (>0.05).

For sensitivity analyses, the prediction performance of the best model LR varied little when stratifying the patient cohort by sex, age, and race/ethnicity (eTable 6-8 in the Supplement). By using different predictors: a) using only ICD codes, the model reached an accuracy of 0.74 for one-year prediction, but failed to make meaningful predictions for ten-year prediction as it labelled all the test patients as AD patients; b) using only structured data (demographics, medications and ICD codes), the model reached an accuracy of 0.79 for one-year prediction and 0.47 for ten-year prediction; c) using structured data with the proposed keywords predictors, the model performed similarly to the keywords-only approach (eTable 9 in the Supplement). Excluding cognitive test relevant predictors such as "MMSE" and "Mini-Cog" did not affect the model performance significantly (eTable 10 in the Supplement). For model interpretation, the most important keywords by Shapley value were "concentration", "language", "speaking", "affect", and "judgement" (eFigure 7 in the Supplement). Using only primary notes reduced the model performance to an accuracy of 0.87 (eTable 11 in the Supplement). We also tested the model on different groups of AD patients based on their diagnosis codes. For ten-year prediction, we had 16 AD cases with a single AD diagnosis at dementia clinics and 1096 AD cases with multiple AD diagnoses in the test set. The model achieved an F1 score of 0.76 on the former group and 0.93 on the latter group. On another group of patients with a single AD diagnosis or no diagnosis from specialized clinics, we collected 700 such patients from the VHA system who had over 15 years of continuous EHR. The model achieved an F1 score of 0.48 for ten-year prediction and 0.66 for one-year prediction. The time interval between the first and second diagnosis for AD case patients with multiple AD diagnoses did not significantly affect the model's performance (eFigure 8 in the Supplement).

Although the predictive model showed excellent performance, it made some incorrect predictions. For the error analysis, out of the total predictions of the LR model, we observed three false positives, where the model predicted high probability of AD (>0.5) but the patient has no clinical AD diagnosis, and conversely 148 are false negative predictions. Domain experts chart-reviewed all three false positive patients and 13 randomly selected false negative patients. They concluded that one of the three false positive patients had mild cognitive impairment (MCI), and the other two had major depressive disorder (MDD). Domain experts found three main features among the 13 false negative patients: 1) misdiagnosed: the patients had non-AD dementia (four patients); 2) insufficient data to make prediction (three patients); 3) machine learning system failures (six patients) (eMethods in the Supplement).

## Discussion

With the global population aging, the prevalence of AD is rapidly increasing.[2] Therapies to prevent or treat early AD appear increasingly important.[33] Effective use of these therapies, though, will depend on the efficient identification of high-risk individuals who might benefit best from therapies.[34] Some efforts have identified predictive factors for AD progression[35] and some individual early identification tools have been developed[36], but to date no efficient approach exists for large-scale screening of patients at high risk of AD. Here we explore whether routine information recorded in EHRs can help with this task.

We found that keywords that may be signs or symptoms of AD, such as forgetfulness or depression, are routinely recorded in the EHRs of our studied patients. These keywords reflect a wide spectrum of physical and psychological conditions noted in routine patient heath management. They are derived from symptoms used for screening individuals for signs of cognitive decline. In addition, some of our keywords were cognitive tests that clinicians might order or perform such as the MMSE. Such testing, performed outside of routine clinical care, may also reflect clinicians recognizing something aberrant in a patient's clinical presentation.

Starting about 14 years prior to the appearance of AD diagnosis recorded in ICD codes, the frequency of occurrence of these keywords began to dramatically increase in EHRs when compared to controls without AD. Thus, patients were mentioning of underlying signs and symptoms, and clinicians believed these mentions were significant enough to record them in their notes, well before there were sufficient changes to diagnose AD. This is consistent with a growing body of knowledge that has shown that biological changes in the brain which happen long before one is diagnosed as AD have noticeable and consistent impacts on one's cognition/behaviors/mood/etc. [19–21] These changes, when observed over an extended period, show strong correlations with AD risk.

Importantly, we found that the occurrence, and pattern, of these keywords is predictive of the development of AD. We derived models using data from at least five years of EHRs, and used varying time intervals between these notes and the diagnosis of AD. This resulted in derivation and validation samples of different sizes as we varied the length of the time period. The LR model consistently performed well in validation samples for making predictions zero to ten years prior to diagnosis with over 90% accuracy rate. The performance was also validated using hold-out VHA stations and by age, sex, and race/ethnicity, suggesting the effectiveness and generalizability of the model. It was also observed that the model's performance remained strong even when cognitive test-related keywords were excluded, indicating that AD-related signs and symptoms identified through routine clinical care records are sufficient for AD risk prediction.

Comparing with making prediction using structured data from EHRs (diagnosis codes and medications), keywords predictors appear to be more representative and indicative of AD risk, as it allows for a wider net to capture AD signs and symptoms that change in frequency.

Early detection of AD is crucial to reduce underdiagnosis, especially in the primary care setting, and to facilitate timely referral and treatment for high-risk patients.[2] Although it is a challenging task, our model shows promise for early screening based on information available in primary care (accuracy 0.87 for prediction ten years prior to diagnosis)).

Our prediction performance for patients under age 65 was poor (accuracy 0.76 for ten-year prediction). This could be due to the low prevalence of this group in our cohort (about 1%). Furthermore, our predictors were the most frequent 1000 keyword-age pairs selected from the whole training data where the majority of patients were over the age of 65.

Therefore, the predictors for this group were sparse which led the model hard to capture the early-onset AD patterns. Previous studies have also indicated that early-onset AD has distinct features from late-onset AD, such as different phenotypic presentations, higher genetic susceptibility, and differences in neuropathologic burden and topography.[25]

We found that the predictive model performed better for patients who had multiple ICD diagnoses of AD in the longitudinal EHR. The model's accuracy was lower for patients who received a single AD diagnosis at dementia clinics, and even lower for those who received one diagnosis at any clinics or two or more at non-specialty clinics. We examined the 279 patients with a single dementia clinic-based AD diagnosis and found that 261 (94%) of them had other diagnoses of dementia or related conditions, such as gait problem, Parkinson's disease, psychosocial problem, MDD, MCI, and post-traumatic stress disorder, after their AD diagnosis. This suggests a high risk of misdiagnosis in this group. The remaining 18 patients (6%) died within 2 years and had insufficient follow-up data for analysis.

Our error analyses show that two of the total three false positives were diagnosed with MDD. This is consistent with existing research that MDD and AD share similar symptoms and biologic pathways.[37] Four (30%) false negative patients were found to have dementia of different types (Lewy body dementia, frontotemporal dementia etc.) by domain experts' chart-review, suggesting the chance of misdiagnosis could be high for these four patients and the challenges for clinical diagnosis of AD.[38]

This study, based on a large national EHR data from the VHA which includes 1,298 health care settings across US, uses one of the largest patient cohorts for AD research with 16,701 AD patients. Over seven million EHR notes are analyzed in this work. The results showed a high performance in making early AD prediction. The greater than 90% accuracy rate compares favorably with other biomarkers such as PET imaging and plasma biomarkers. Comparing with recent works for prediction employing predictors like grip strength, reaction time, prospective memory, cognitive testing etc.,[21] which are not quite as easy to collect as keywords in regular health care visits, the proposed keyword screen is more efficient with machine learning prediction filling in gaps in testing accessibility and affordability. It could potentially be used to examine a large volume of clinical notes for screening populations at risk for AD.

**Limitations**

Our study has several limitations that should be considered. First, we used VHA data, which may not represent the general population, as VHA patients tend to be demographically skewed, socioeconomically disadvantaged, and have a high prevalence of conditions such as post-traumatic stress disorder and traumatic brain injury. Our findings need to be validated on non-VHA data in future work. Second, we conducted a case-control study, which is simpler than a cohort study, but less able to show causality and more susceptible to bias.[39] A cohort study was not feasible for this work due to the computational cost of analyzing over 15 years of longitudinal data. Moreover, our study relied on the assumption that EHR data were accurate and complete in capturing historical symptoms and AD diagnosis. However, EHR data have inherent limitations,[40,41] such as the potential for confounding bias due to a lack of randomization and for selection bias due to missing data, as well as AD diagnostic challenges. One way we tried to address these limitations was performing sensitivity analysis on different groups of patients with different diagnosis codes. Third, we excluded patients with less than 5 years of continuous EHR history, which may have introduced selection bias, as patients with lower hospital utilization and less health-conscious clinical visits may benefit more from large-scale screening, but were underrepresented in our study. We will explore additional data sources to include this group of patients in our prediction model. Fourth, due to the low prevalence of patients under age 65 and female patients in the cohort, we did not examine the specific AD patterns for these two subgroups for optimal performance. We plan to conduct separate studies to better understand these subgroups of patients, and also explore methods that handle imbalanced data for better prediction performance on those patients. Fifth, our study period did not include any effective interventions to halt or cure early AD symptoms, which may have affected the trends we observed for AD signs and symptoms. The availability and adoption of such interventions may change the natural history and progression of AD, and make prediction more challenging. We will monitor the impact of new therapies on our approach. Sixth, our study focused on screening for AD in the general population, and did not specifically address non-AD type dementia, which may have different clinical features and outcomes. Although we did not explicitly exclude non-AD dementia patients from the control cohort, they would only constitute a small proportion of the general non-AD population. We will evaluate the efficacy of our approach in distinguishing between AD and other types of dementia in future work. Seventh, the AD-related keywords we used might not accurately capture AD-related signs and symptoms for the lack of clinical context. In future studies, we will explore a more precise method to identify AD-related signs and symptoms for future work.

## Conclusions

Signs and symptoms suggestive of AD start appearing with increasing frequency in clinicians' notes as early as 14 years before the diagnosis is made. These may be useful to help identify people at future risk for AD.

## Acknowledgment

The authors have no conflicts of interest. Research reported in this study was supported by the National Institute on Aging of the National Institutes of Health (NIH) under award number R01AG080670. The content is solely the responsibility of the authors and does not necessarily represent the official views of the NIH.

**Figure 1. AD case-control cohort creation flowchart**

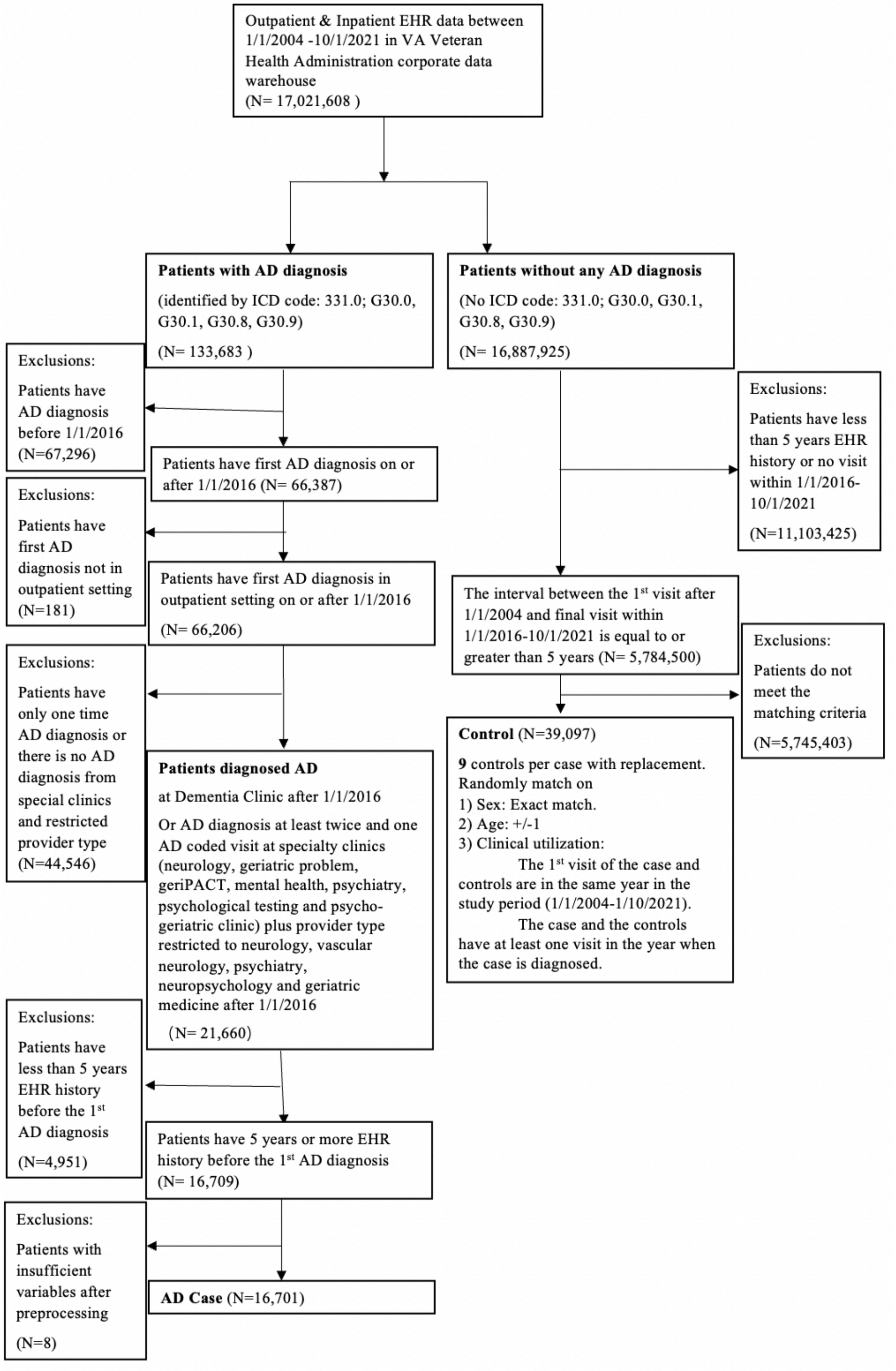

**Figure 2. Method workflow: from extracting AD-relevant keywords from longitudinal EHRs to predictions for decision support**

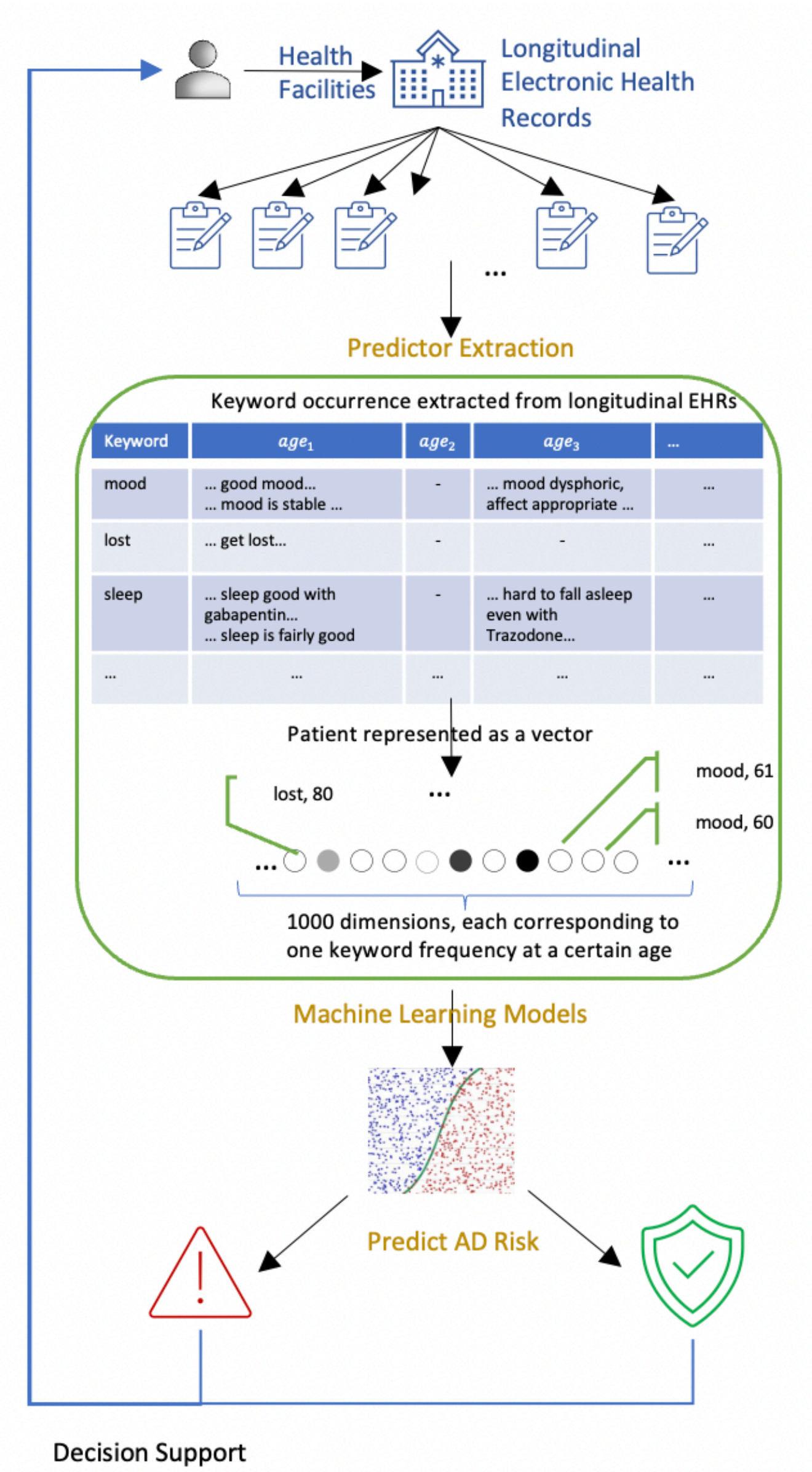

The framed part shows the predictors and their contributions to the prediction.

**Figure 3: Study timeline and prediction window setting**

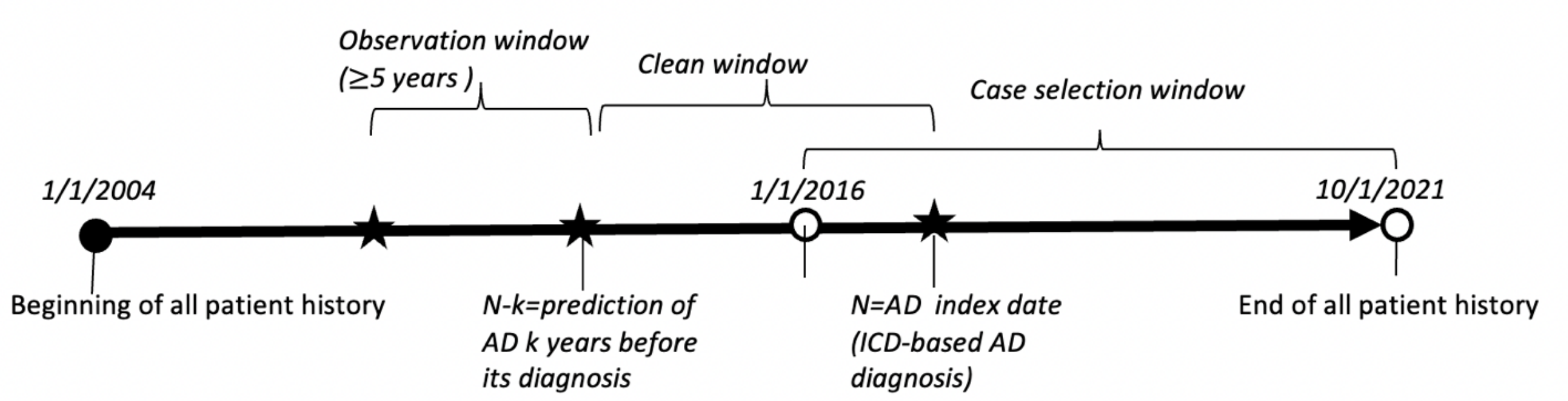

The study period is from 1/1/2004 to 10/1/2021 and AD index date is after 1/1/2016. At least 5 years of longitudinal EHRs are used at the time of prediction. Clean period during which no data are used in predicting AD development could vary from 0 to 10 years.

**Figure 4. Plots of the average number of AD-related keyword counts per patient by year/age**

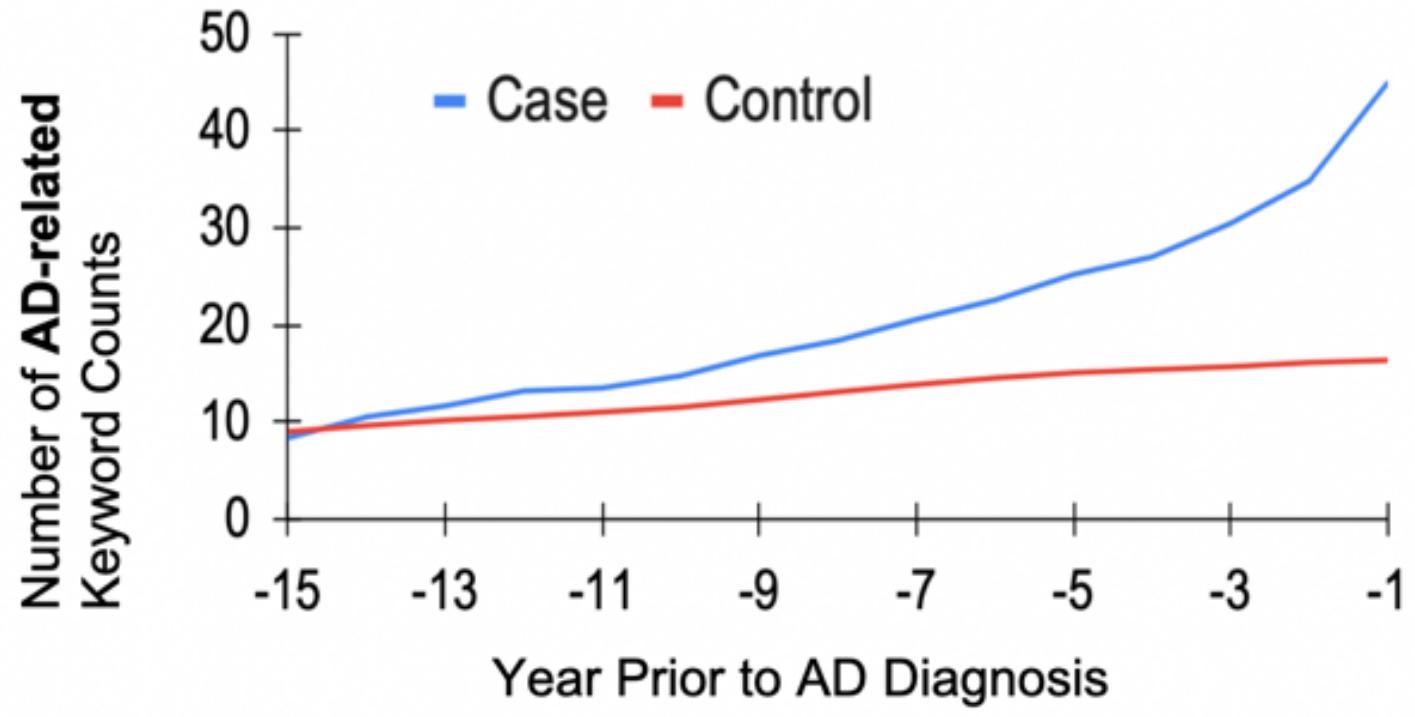

(a)

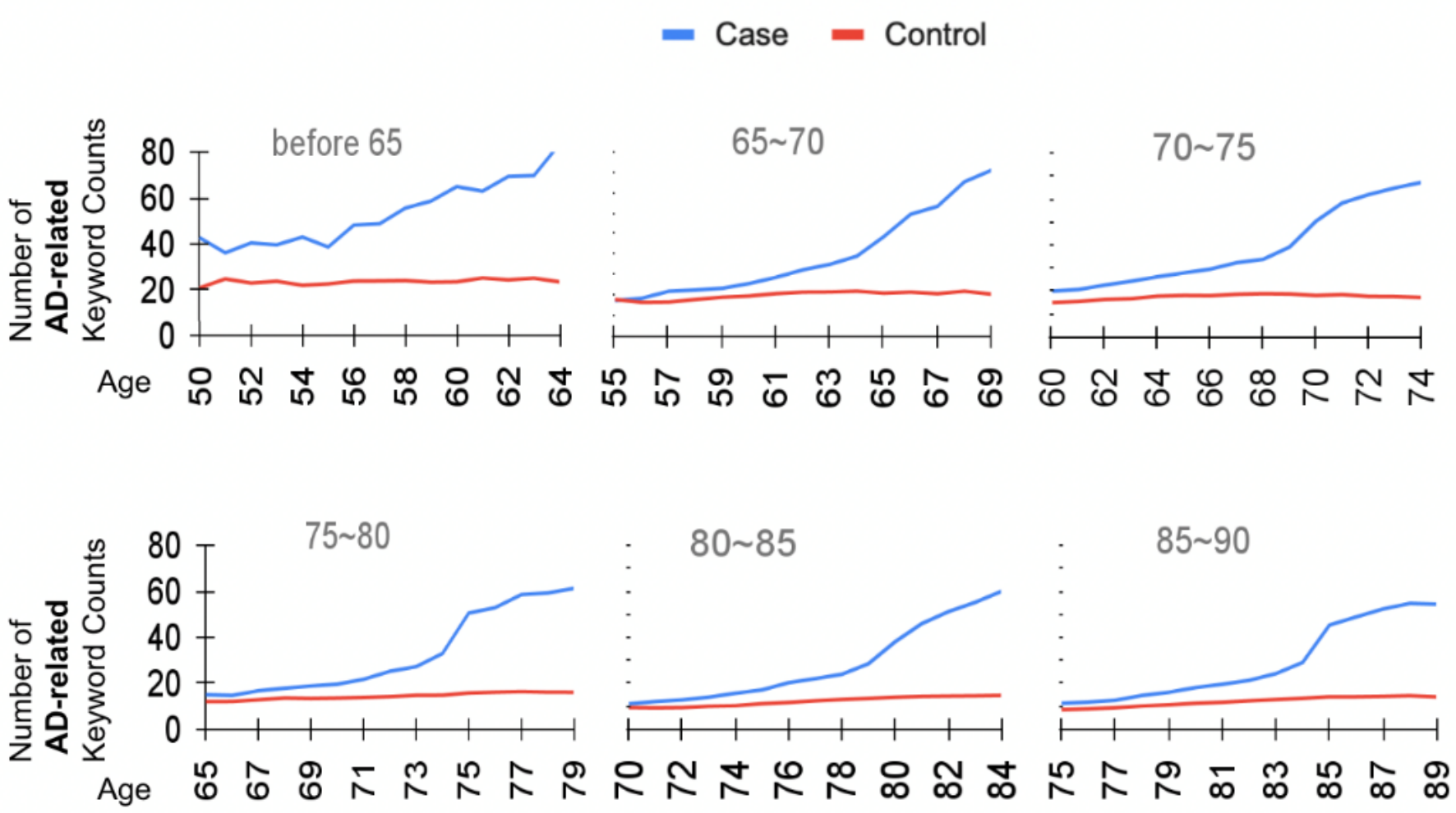

(b)

(a). AD case patients with matched control patients: plots of the average number of AD-related keyword counts by year prior to AD diagnosis. (b). AD case patients diagnosed at specific age groups (<65, 65-70, 70-75, 75-80, 80-85, 85-90) with matched control patients: plots of the average number of AD-related keyword counts by age. The statistics are derived from the entire cohort (AD cases: 16701; Control: 39097). The expert-curated AD-related keywords are used for identifying AD signs and symptoms.

**Table 1. Characteristics of the study cohort**

| Characteristic | AD Case (n=16,701) | Control (n=39,097) |
|---|---|---|
| Unique patients, No. | 16,701 | 39,097 |
| Age, Mean (SD), y | 76.7 (8.6) | 76.6 (8.9) |
| Sex/Gender, No. (%) | | |
| Female | 488 (2.9%) | 1,013 (2.6%) |
| Male | 16,213 (97.1%) | 38,084 (97.4%) |
| Race/Ethnicity, No. (%) | | |
| White | 12,927 (77.4%) | 31,256 (79.9%) |
| Black or African American | 2,356 (14.1%) | 4,024 (10.3%) |
| Asian | 121 (0.7%) | 260 (0.7%) |
| Native Hawaiian or Other Pacific Islander | 168 (1.0%) | 297 (0.8%) |
| American Indian or Alaska Native | 86 (0.5%) | 225 (0.6%) |
| Unknown | 1,043 (6.2%) | 3,035 (7.8%) |
| Hispanic or Latino | 1,865 (11.2%) | 1,759 (4.5%) |
| Charlson Comorbidity Index, Mean (SD) | 3.74 (3.11) | 2.92 (2.91) |
| | (Before AD onset) | (Across the period) |
| | | |

| **Characteristic** | **AD Case (n=16,701)** | **Control (n=39,097)** |
| --- | --- | --- |
| Number of notes with keywords (average) | 92 | 77 |
| Keywords of Interests number (average) | 532 | 214 |

**Table 2. Ten years prior to ICD-based diagnosis prediction results**

| Model | AD Case (P/R/F) | Control (P/R/F) | Accuracy | PR AUC | ROC AUC |
|---|---|---|---|---|---|
| Setting I (on a randomly sampled test set)<br>Total patients No. in test set: 4076, AD cases No.: 1112 | | | | | |
| LR | 1.00/0.87/0.93 | 0.95/1.00/0.98 | 0.96 | 0.990 | 0.997 |
| SVM | 0.99/0.83/0.91 | 0.94/1.00/0.97 | 0.95 | 0.985 | 0.995 |
| AdaBoost | 1.00/0.60/0.75 | 0.87/1.00/0.93 | 0.89 | 0.862 | 0.960 |
| RF | 1.00/0.32/0.48 | 0.80/1.00/0.89 | 0.81 | 0.885 | 0.926 |
| Setting II (on a test set built from patients of 10 hold-out stations)<br>Total patients No. in test set: 1670, AD cases No.: 493 | | | | | |
| LR | 1.00/0.86/0.93 | 0.95/1.00/0.97 | 0.96 | 0.998 | 0.991 |
| SVM | 1.00/0.85/0.92 | 0.94/1.00/0.97 | 0.96 | 0.997 | 0.991 |
| AdaBoost | 1.00/0.60/0.75 | 0.86/1.00/0.92 | 0.88 | 0.95 | 0.851 |
| RF | 1.00/0.34/0.51 | 0.78/1.00/0.88 | 0.80 | 0.946 | 0.896 |

[a]LR: logistic regression.

[b]SVM: support vector machine.

[c]RF: random forests.

[d]P/R/F: precision/recall/F1 score.

[e]PR AUC: area under the precision-recall curve.

[f]ROC AUC: area under the receiver operating characteristic curve.

.